\documentclass[10pt,twocolumn,letterpaper]{article}

\usepackage{iccv}
\usepackage{times}
\usepackage{epsfig}
\usepackage{graphicx}
\usepackage{amsmath}
\usepackage{amssymb}
\usepackage{verbatim}
\usepackage{multirow}
\usepackage{subfig}
\usepackage{theorem}
{\theorembodyfont{\rmfamily} }

\usepackage[pagebackref=true,breaklinks=true,letterpaper=true,colorlinks,bookmarks=false]{hyperref}

\iccvfinalcopy 

\def\iccvPaperID{9} 

\ificcvfinal\fi
\begin{document}

\title{Standard detectors aren't (currently) fooled by physical adversarial stop signs}

\author{Jiajun Lu, Hussein Sibai, Evan Fabry, David Forsyth\\
University of Illinois at Urbana Champaign\\
\{jlu23, sibai2, efabry2, daf\}@illinois.edu
}

\maketitle

\begin{abstract}
An {\em adversarial example} is an example that has been adjusted to produce the wrong label when presented to a system at test  time.  If adversarial examples existed that
could fool a detector, they could be used to (for example) wreak havoc on roads populated with smart vehicles. 
Recently, we described our difficulties creating physical adversarial stop signs that fool a detector.  More recently, Evtimov {\em et al.} produced a physical adversarial stop sign that fools a proxy model of a detector.  
In this paper, we show that these physical adversarial stop signs do not fool two standard detectors (YOLO and Faster RCNN) in standard configuration.   Evtimov {\em et al.}'s construction relies on a crop of
the image to the stop sign; this crop is then resized and presented to
a classifier.   We argue that the cropping and resizing procedure
largely eliminates the effects of rescaling and of view angle.
Whether an adversarial attack is robust under rescaling and change of
view direction remains moot.  We argue that attacking a classifier is
very different from attacking a detector, and that the structure of
detectors -- which must search for their own  bounding box, and which
cannot estimate that box very accurately -- likely  makes it difficult to 
make adversarial patterns.    Finally, an adversarial pattern on a physical object that could fool a detector would have
to be adversarial in the face of a wide family of parametric distortions (scale; view angle; box shift inside the
detector; illumination; and so  on).   Such a pattern would be of great theoretical and practical interest.  There is
currently no evidence that such patterns exist.  
\end{abstract}

\section{Introduction}

An {\em adversarial example} is an example that has been adjusted to produce the wrong label when presented to a system
at test  time. Adversarial examples are of interest only because the adjustments required seem to be very small and are
easy to obtain~\cite{szegedy2013intriguing, goodfellow2014explaining, DBLP:journals/corr/FawziMF16}.  Numerous search
procedures generate adversarial examples~\cite{DBLP:journals/corr/Moosavi-Dezfooli15, nguyenfooled,
  moosavi2016universal}.  There is fair evidence that it is hard to tell whether an example is adversarial (and so (a)
evidence of an attack and (b) likely to be misclassified) or not~\cite{shaham2015understanding,
  DBLP:journals/corr/GuR14, Sharif:2016:ACR:2976749.2978392, metzen2017detecting, 
  carlini2016defensive, fawzi2015analysis}.  Current procedures to build adversarial examples for deep networks appear
to subvert the feature construction implemented by the network to produce odd patterns of activation in late stage
RELU's; this can be exploited to build one form of defence~\cite{lu2017safetynet}.   There is some evidence that other
feature constructions admit adversarial attacks, too~\cite{metzen2017detecting}.  The success of these attacks can be
seen as a warning not to use very highly non-linear feature constructions without  having strong mathematical
constraints on what these constructions can do; but taking that position means one cannot use methods that are largely
accurate and effective. 

It is important to distinguish between a classifier and a detector to understand the current state of the art.  A
classifier accepts an image and produces a label. Classifiers are scored on accuracy. A detector, like
FasterRCNN~\cite{ren2015faster}, identifies image boxes that are ``worth labelling'', and
then generates labels (which might include {\tt background}) for each.   The final label generation step employs a
classifier.  However, the statistics of how boxes span objects in a detector are complex and poorly understood.  Some
modern detectors like YOLO 9000~\cite{redmon2016yolo9000} predict boxes and labels using features on a fixed grid, resulting in fairly complex sampling
patterns in the space of boxes, and meaning that pixels outside a box may participate in labelling that box.   One
cannot have too many boxes, because too many boxes means much redundancy; worse, it imposes heavy demands on the
accuracy of the classifier.  Too few boxes chances missing objects.  Detectors are scored on a composite score, taking
into account both the accuracy with which the detector labels the box and the accuracy of the placement of the box.  

It is usual to attack classifiers, and all the attacks of which we are aware are attacks on classifiers.  However, for
many applications, classifiers are not themselves useful.  Road signs are a good example.  A road sign classifier would
be applied to images that consist largely of road sign (e.g. those of~\cite{stallkamp2012man}).  But there is little application need for a road-sign classifier except as a
component of a road sign detector, because one doesn't usually have to deal with images that consist largely of road
sign.  Instead, one deals with images that contain many things, and must find and label the road sign. It is quite
natural to study road sign classifiers (e.g.~\cite{sermanet2011traffic}) because image classification remains difficult and academic
studies of feature constructions are important.  But there is no particular threat posed by an attack on a road sign
classifier.  An attack on a road sign detector is an entirely different matter.  For example, imagine possessing a
template that, with a can of spray paint, could ensure that a detector read a stop sign as a yield sign (or worse!). As
a result, it is important to know whether (a) such examples could exist and (b) how robust their adversarial property is in practice.

Printing adversarial images then photographing them can retain their adversarial
property~\cite{DBLP:journals/corr/KurakinGB16, athalye2017synthesizing}, which suggests adversarial examples might exist
in the physical world.  Adversarial examples in the physical world could cause a great deal of mischief.    In earlier
work, we showed that it was difficult to build physical examples that fooled a stop-sign  detector~\cite{lu2017no}.  In
particular, if one actually takes video of adversarial stop-signs out of doors, the adversarial pattern  does not
appear to affect the performance of the detector by much.  We speculated that this might be because adversarial
patterns were disrupted by being viewed at different scales, rotations, and orientations.  This generated some
discussion.  OpenAI demonstrated a search procedure that could produce an image of a cat that was misclassified  when
viewed at multiple scales~\cite{athalye2017synthesizing}.  There is some blurring of the fur texture on the  cat, but
this would likely be imperceptible to most observers.  OpenAI also demonstrated a search procedure that could  produce
an image of a cat that was misclassified when viewed at multiple scales {\em and}
orientations~\cite{athalye2017synthesizing}.  However, there are significant visible artifacts on that image; few would
feel that it had not  obviously been tampered with.

Recently, Evtimov {\em et al.} have demonstrated several physical stop-signs that are
misclassified~\cite{evtimov2017robust}.   Their attack is demonstrated on stop-signs that are cropped from images and
presented to a classifier.  By cropping, they have proxied the box-prediction process in a detector; however, their
attack is not intended as an attack on a detector (the paper does not use the word ``detector'', for example).  In this
paper, we show that standard off-the-shelf detectors that have not seen adversarial examples in training detect their
stop signs rather well, under a variety of conditions.   We explain (a) why their result is puzzling; (b) why their
result may have to do with specific details of their pipeline model, particularly the classifier construction and (c)
why the distinction between a classifier and a detector means their work has  not put the  core issue -- can one build
physical adversarial stop-signs? -- to rest.  

\section{Experimental Results}

Evtimov {\em et. al} have demonstrated a construction of physical adversarial stop signs~\cite{evtimov2017robust}.  They demonstrate poster attacks (the stop sign is covered with a poster that looks like a faded stop sign) and sticker attacks (the attacker makes stickers placed on particular locations on a stop sign), and conclude that one can make physical adversarial stop signs.  There are two types of tests: stationary tests, where the sign is imaged from a variety of orientations and directions; and drive-by tests, where the sign is viewed from a camera based on a car.

\begin{figure*}[!t]
	\centerline{\includegraphics[width=1.0\linewidth]{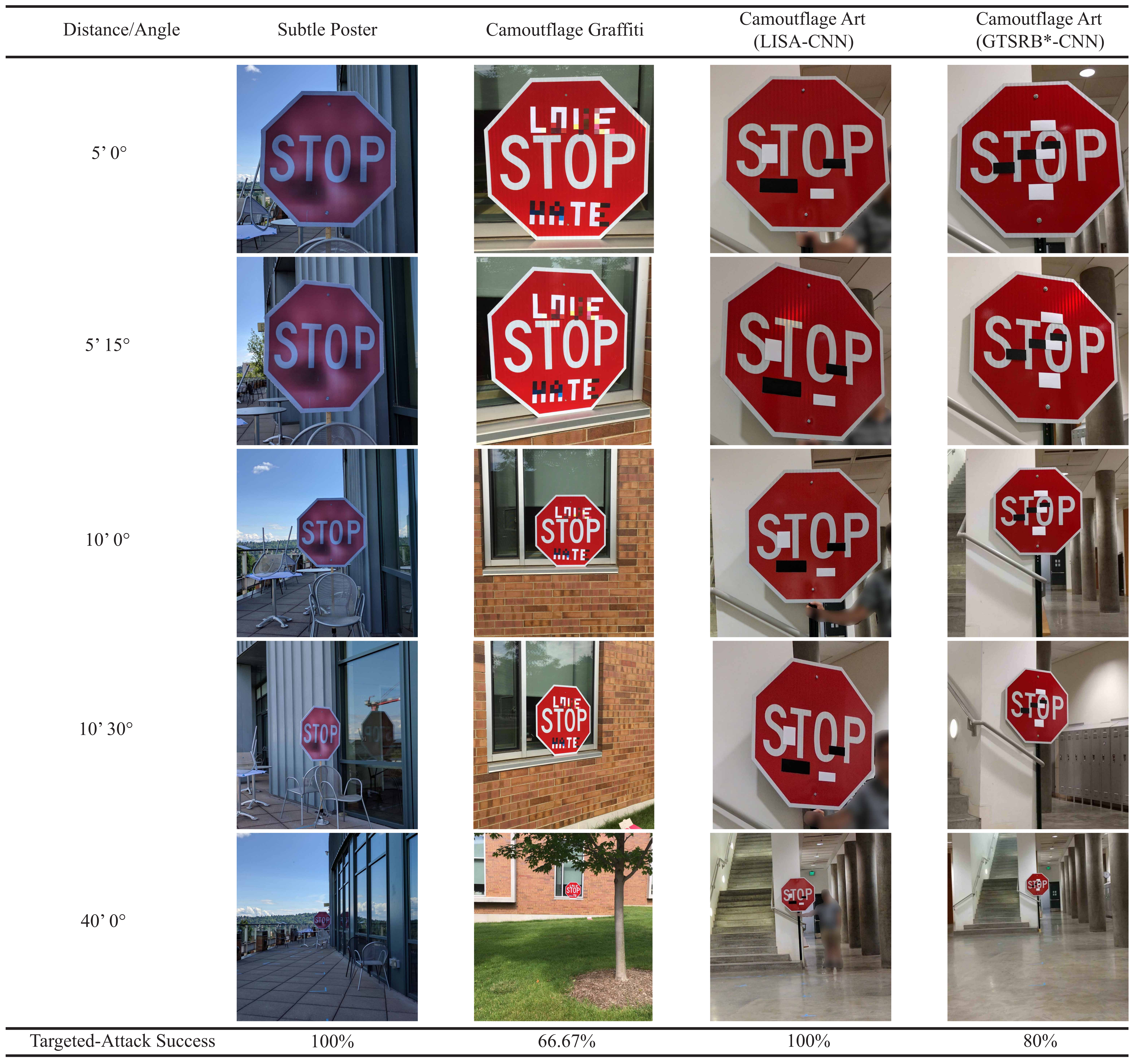}}
	\caption{Table IV of \protect \cite{evtimov2017robust}, reproduced for the readers' convenience.  This table shows figures
  of different adversarial constructions, from different distances and viewed at different angles.}
	\label{fig:tiv}
\end{figure*}

\begin{figure*}[!t]
	\centerline{\includegraphics[width=1.0\linewidth]{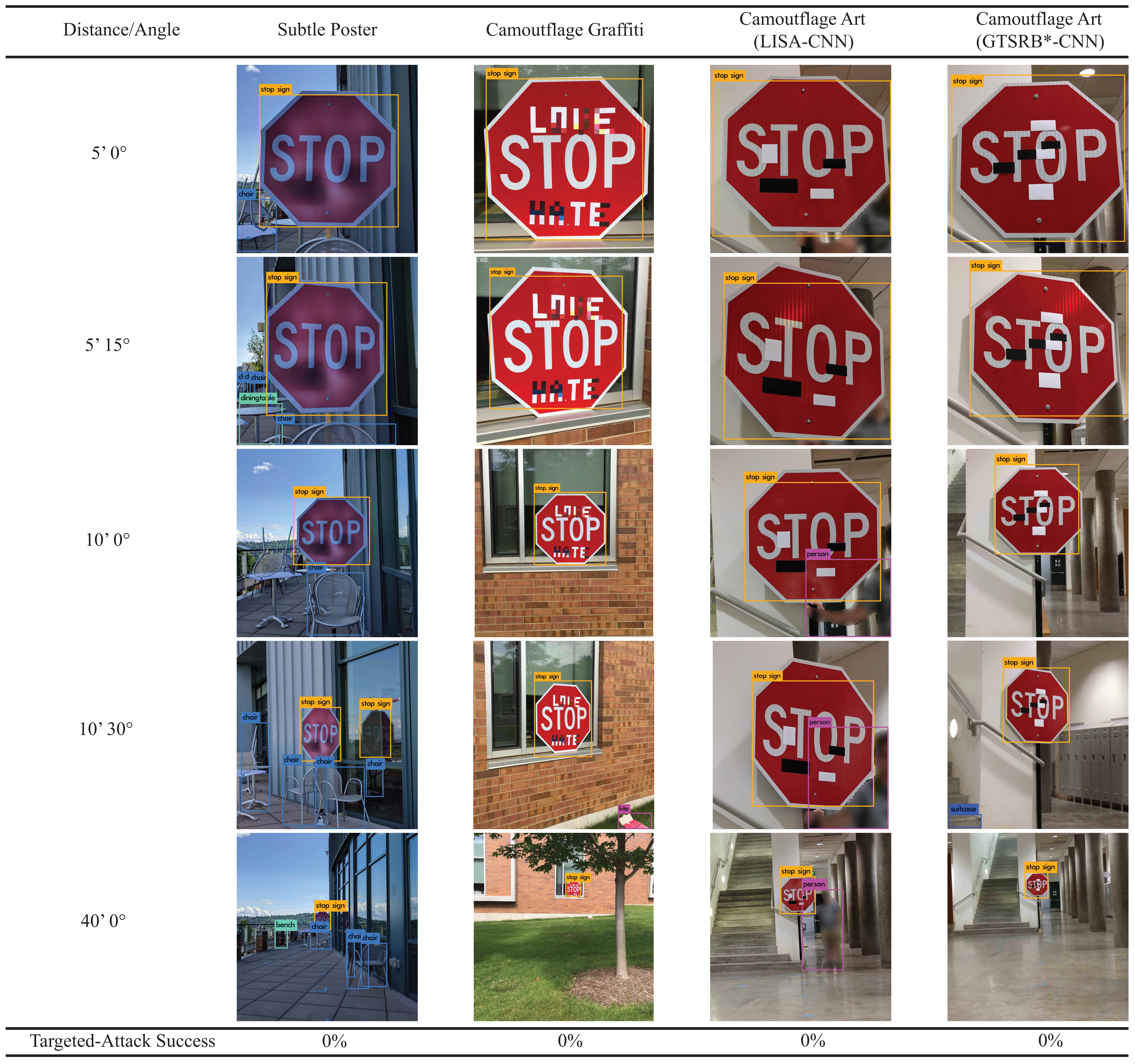}}
	\caption{YOLO detection results on the stop signs of figure \protect \ref{fig:tiv}.}
	\label{fig:detect_imageyolo}
\end{figure*}

\begin{figure*}[!t]
	\centerline{\includegraphics[width=1.0\linewidth]{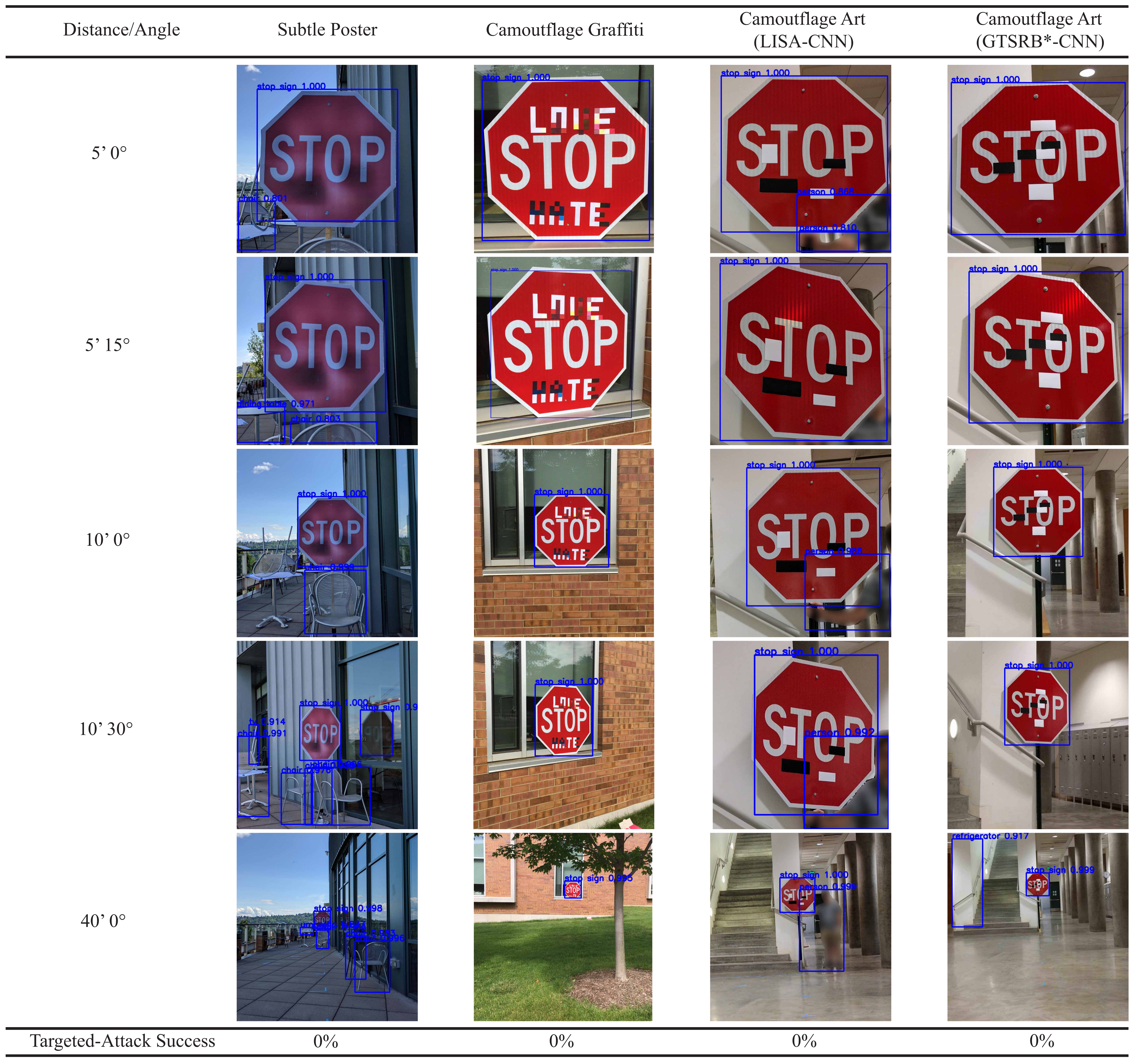}}
	\caption{Faster RCNN detection results on the stop signs of figure \protect \ref{fig:tiv}.}
	\label{fig:detect_imagefrcnn}
\end{figure*}

We obtained two standard detectors (the MS-COCO pretrained standard YOLO~\cite{redmon2016yolo9000}; Faster
RCNN~\cite{ren2015faster}, pretrained version available on github) and applied them to the images and videos from their paper. First, we applied both detectors on
the images shown in the paper (reproduced as Figure~\ref{fig:tiv} for reference).  All adversarial  stop-signs are
detected by both detectors (Figure~\ref{fig:detect_imageyolo} and Figure~\ref{fig:detect_imagefrcnn}).   

\begin{figure*}
	\centerline{\includegraphics[width=1.0\linewidth]{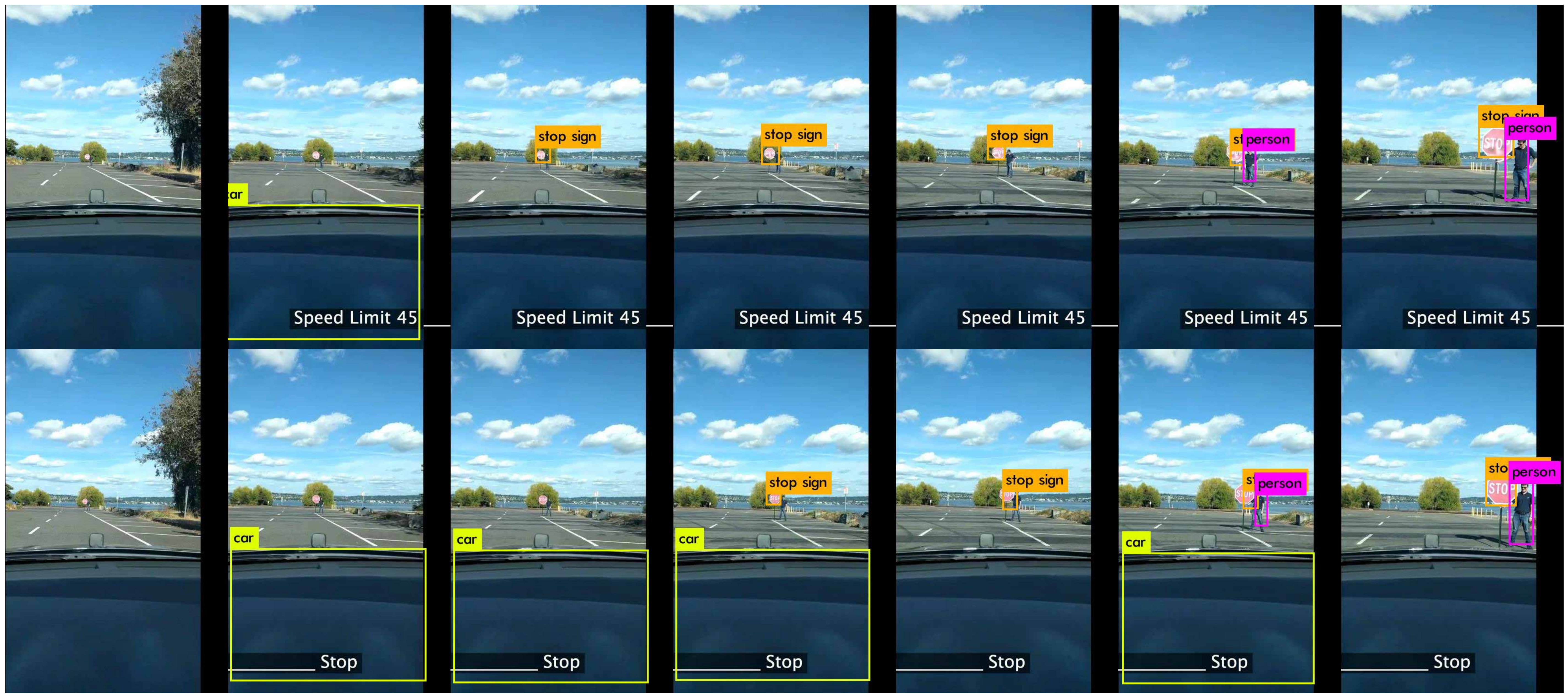}}
	\caption{In relatively low resolution, YOLO detects printed poster physical adversarial stop sign and real stop sign similarly. }
	\label{fig:detect_video4}
\end{figure*}

\begin{figure*}
	\centerline{\includegraphics[width=1.0\linewidth]{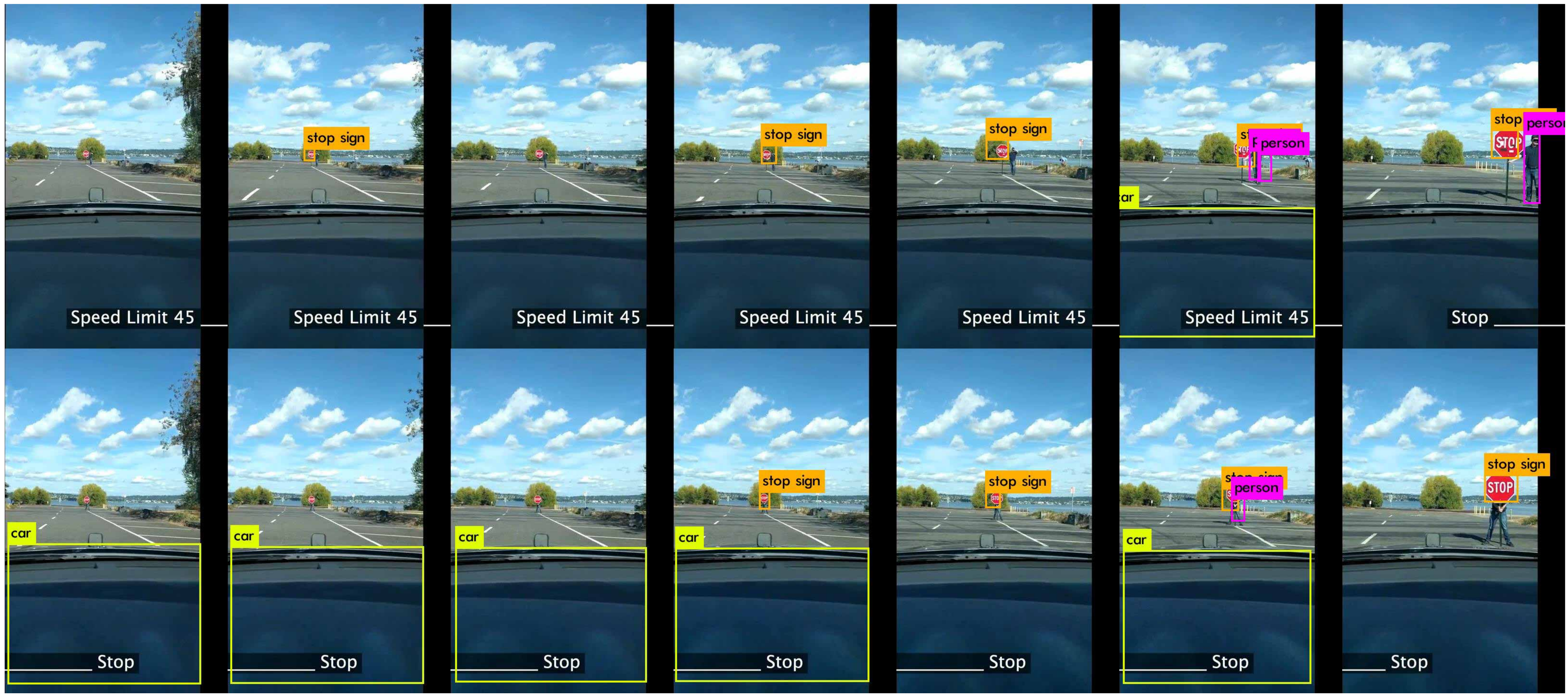}}
	\caption{In relatively low resolution, YOLO detects sticker physical adversarial stop sign and real stop sign similarly. }
	\label{fig:detect_video3}
\end{figure*}

\begin{figure*}
	\centerline{\includegraphics[width=1.0\linewidth]{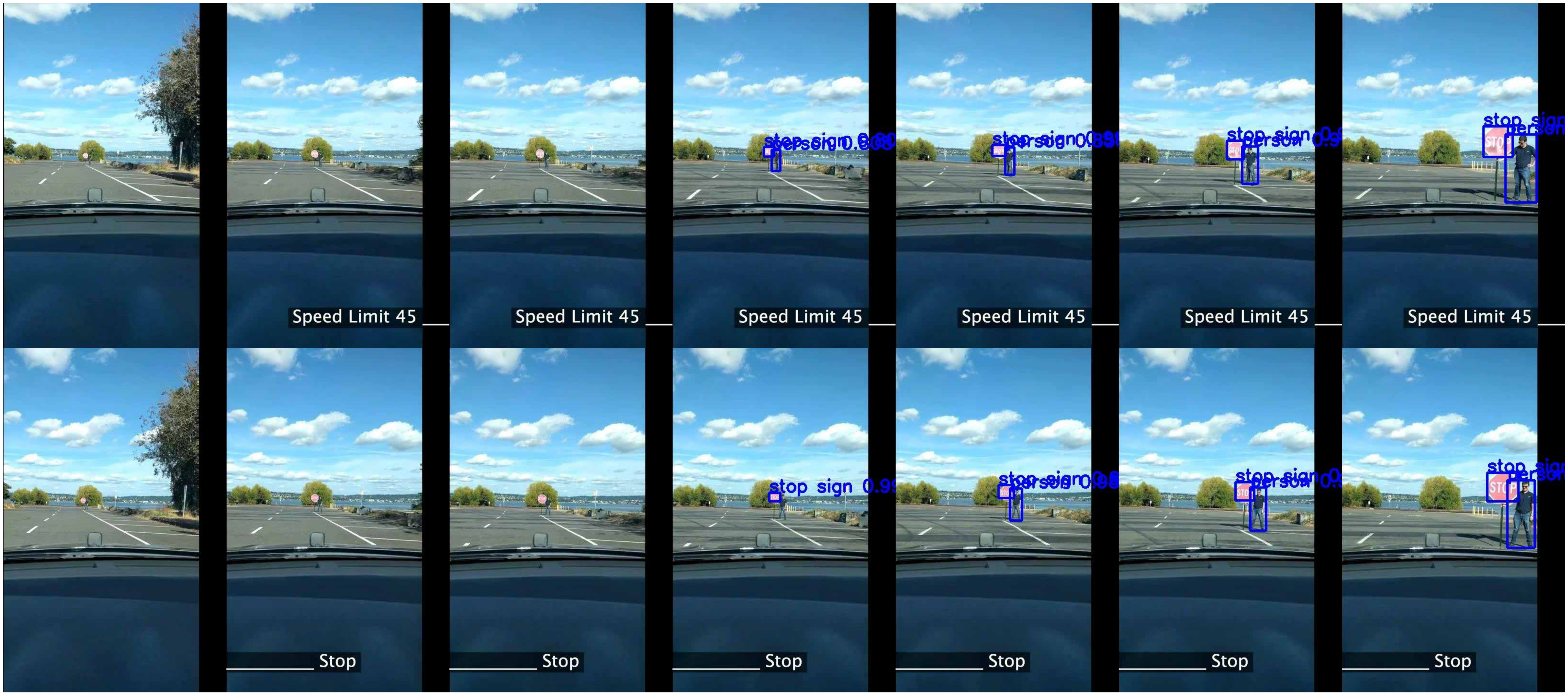}}
	\caption{In relatively low resolution, Faster RCNN detects printed poster physical adversarial stop sign and real stop sign similarly. }
	\label{fig:detect_video6}
\end{figure*}

\begin{figure*}
	\centerline{\includegraphics[width=1.0\linewidth]{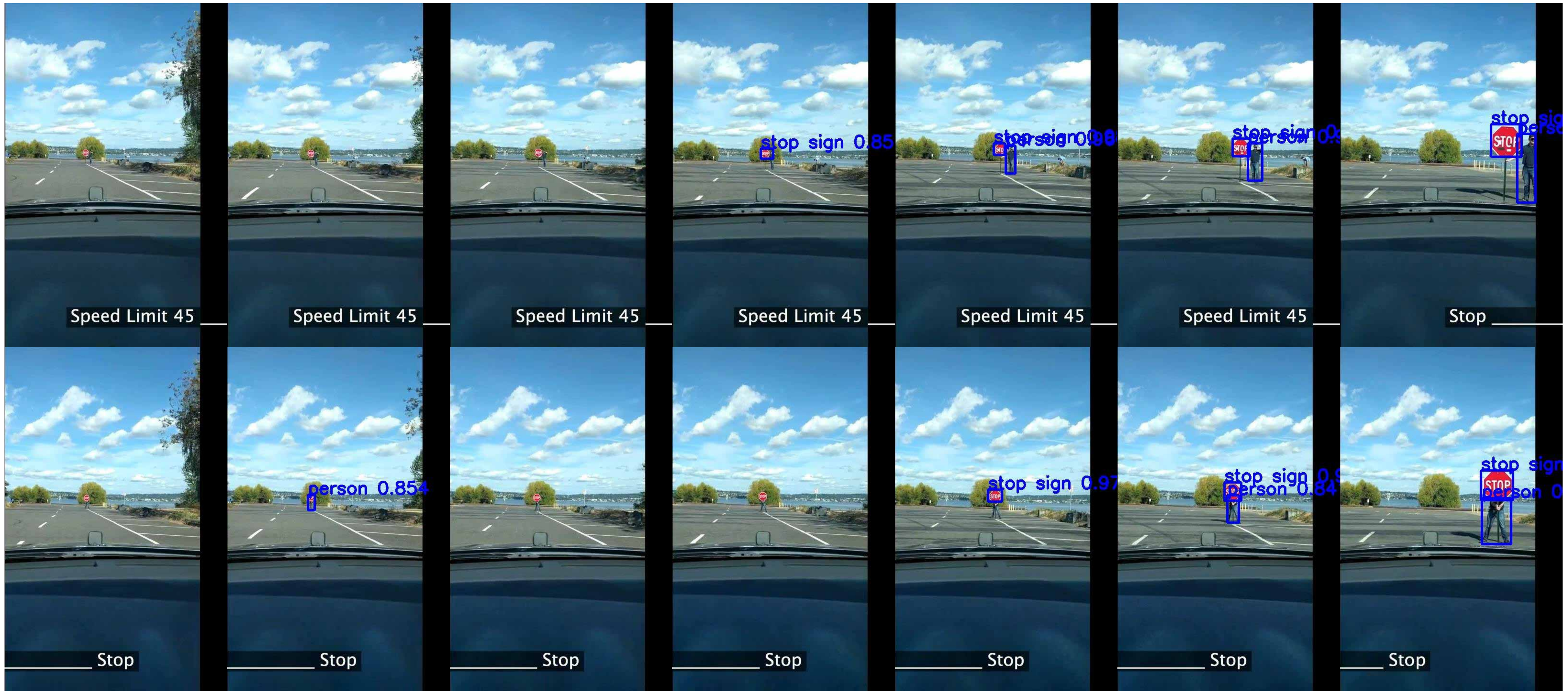}}
	\caption{In relatively low resolution, Faster RCNN detects sticker physical adversarial stop sign and real stop sign similarly. }
	\label{fig:detect_video5}
\end{figure*}

 We downloaded videos provided
by the authors at \url{https://iotsecurity.eecs.umich.edu/#roadsigns}, and applied the detectors to those videos.
We find:
\begin{itemize}
\item  YOLO detects the adversarial stop signs produced by poster attacks about as well as the true stop signs (figure~\ref{fig:detect_video4}, and the videos we provide at \url{https://www.youtube.com/watch?v=EfbonX1lE5s});
\item  YOLO detects the adversarial stop signs produced by sticker attacks about as well as the true stop signs (figure~\ref{fig:detect_video3}, and the videos we provide at \url{https://www.youtube.com/watch?v=GOjNKQtFs64});
\item Faster RCNN  detects the adversarial stop signs produced by poster attacks about as well as the true stop signs (figure~\ref{fig:detect_video6}, and the videos we provide at \url{https://www.youtube.com/watch?v=x53ZUROX1q4});
\item Faster RCNN detects the adversarial stop signs produced by sticker attacks about as well as the true stop signs (figure~\ref{fig:detect_video5}, and the videos we provide at \url{https://www.youtube.com/watch?v=p7wwvWdn2pA});
\item Faster RCNN detects stop signs rather more accurately than YOLO;
\item both YOLO and Faster RCNN detect small stop signs less accurately; as the sign shrinks in the image,
YOLO fails significantly earlier than Faster RCNN.
\end{itemize}
These effects are so strong that there is no point in significance testing, etc.

\pagebreak

Video can be found at:
\begin{itemize}
\item \url{https://www.youtube.com/watch?v=afIr6_cvoqY} (YOLO; poster);
\item \url{https://www.youtube.com/watch?v=rqLhTZZ0U2w}) (YOLO; poster);
\item \url{https://www.youtube.com/watch?v=Ep-aE8T3Igs}  (YOLO; sticker);
\item  \url{https://www.youtube.com/watch?v=nCcoJBQ8C3c} (YOLO; sticker);
\item \url{https://www.youtube.com/watch?v=10DDFs73_6M} (FasterRCNN; poster);
\item \url{https://www.youtube.com/watch?v=KQyzQtuyZxc} (FasterRCNN; poster);
\item \url{https://www.youtube.com/watch?v=FRDyz7tDVdM} (FasterRCNN; sticker);
\item \url{https://www.youtube.com/watch?v=F-iefz8jGQg} (FasterRCNN; sticker).
\end{itemize}

\begin{figure*}
	\centerline{\includegraphics[width=1.0\linewidth]{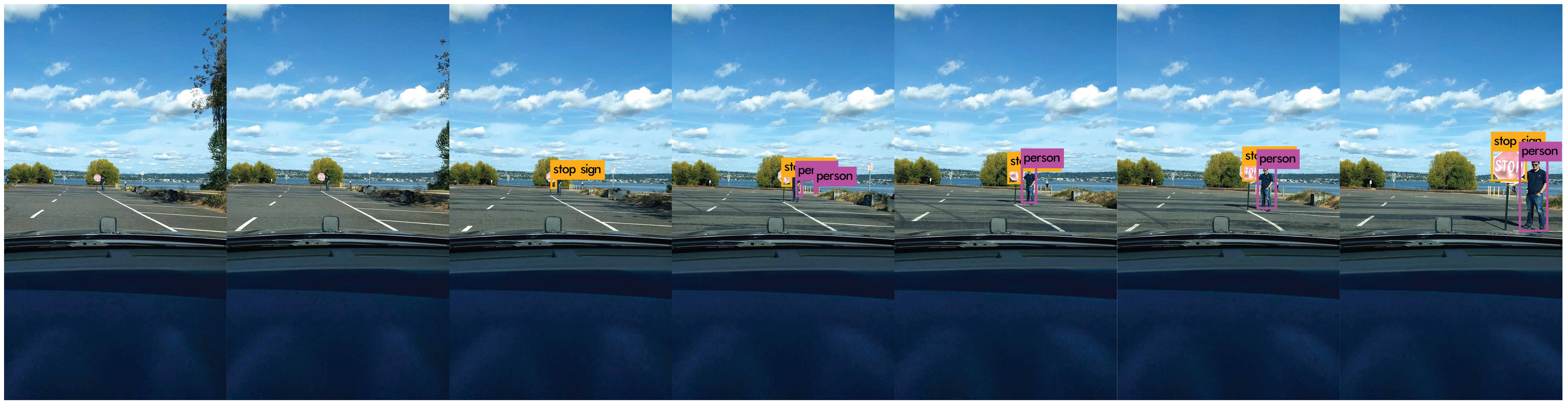}}
	\caption{In higher resolution video, YOLO detects printed poster physical adversarial stop sign well.  YOLO works better on higher resolution than lower resolution video.}
	\label{fig:detect_video_yolo}
\end{figure*}

\begin{figure*}
	\centerline{\includegraphics[width=1.0\linewidth]{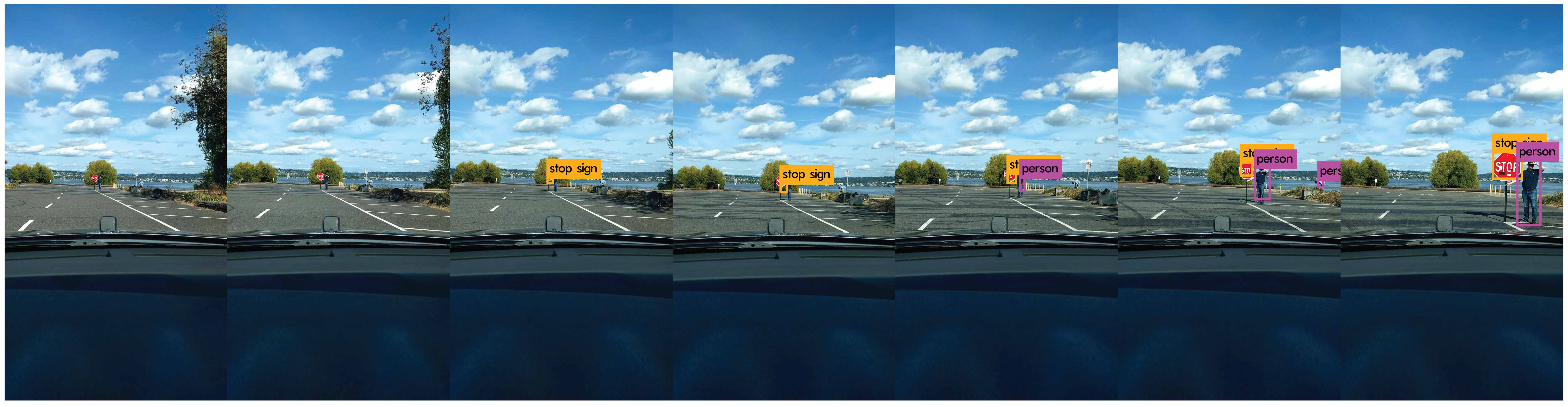}}
	\caption{In higher resolution video, YOLO detects sticker physical adversarial stop sign well. YOLO works better on higher resolution than lower resolution video. }
	\label{fig:detect_video2_yolo}
\end{figure*}

\begin{figure*}
	\centerline{\includegraphics[width=1.0\linewidth]{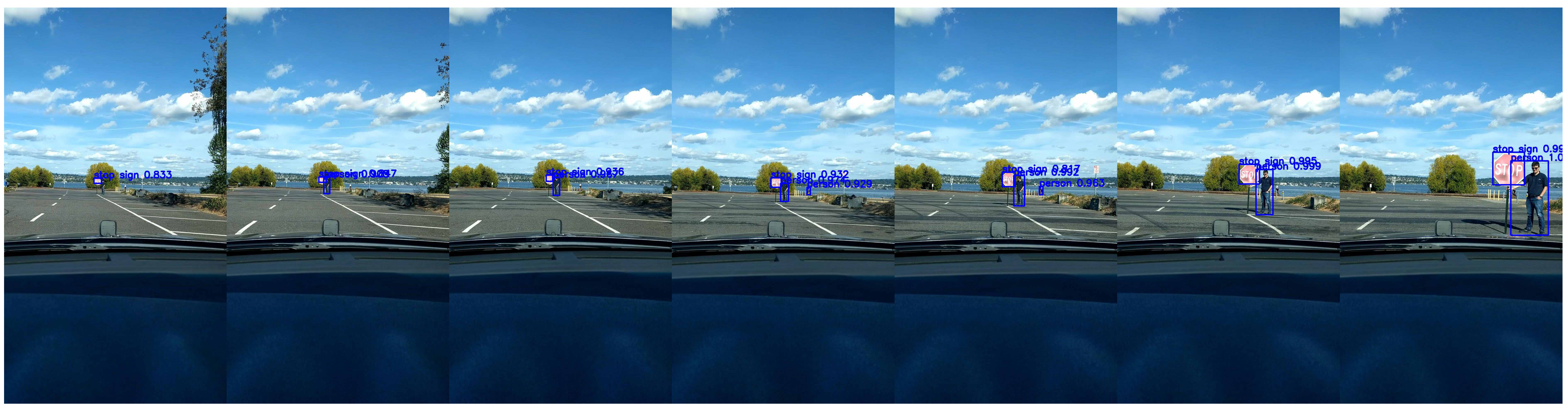}}
	\caption{In higher resolution video, Faster RCNN detects printed poster physical adversarial stop sign very well. }
	\label{fig:detect_video}
\end{figure*}

\begin{figure*}
	\centerline{\includegraphics[width=1.0\linewidth]{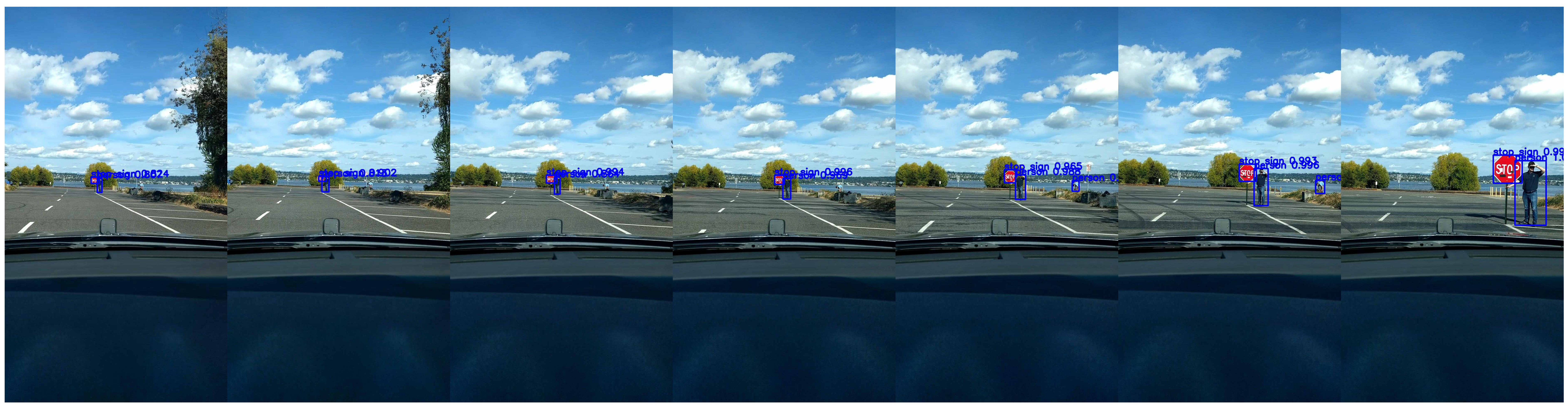}}
	\caption{In higher resolution video, Faster RCNN detects sticker physical adversarial stop sign very well. }
	\label{fig:detect_video2}
\end{figure*}

At our request, the authors kindly provided full resolution versions of the videos at \url{https://iotsecurity.eecs.umich.edu/#roadsigns}.  We applied YOLO and Faster RCNN detectors to those videos.
We find:

\begin{itemize}
\item  YOLO detects the adversarial stop signs produced by poster attacks well (figure~\ref{fig:detect_video_yolo});
\item  YOLO detects the adversarial stop signs produced by sticker attacks (figure~\ref{fig:detect_video2_yolo});
\item Faster RCNN  detects the adversarial stop signs produced by poster attacks very well (figure~\ref{fig:detect_video});
\item Faster RCNN detects the adversarial stop signs produced by sticker attacks very well (figure~\ref{fig:detect_video2});
\item Faster RCNN detects stop signs rather more accurately than YOLO;
\item YOLO works better on higher resolution video;
\item Faster RCNN detect even far and small stop signs accurately.
\end{itemize}

These effects are so strong that there is no point in significance testing, etc.

\begin{figure*}
	\centerline{\includegraphics[width=1.0\linewidth]{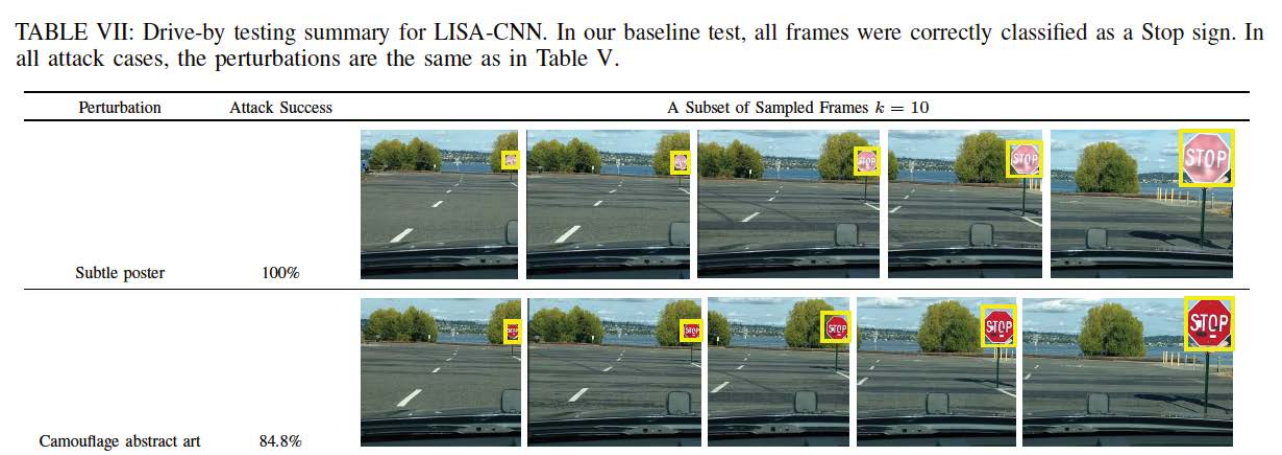}}
	\caption{Table VII of \protect \cite{evtimov2017robust}, reproduced for the reader's convenience.  Note the
          close crops of stop signs, shown as yellow boxes.    The whole image could not be passed to a stop sign {\em
            classifier}; therefore, some form of box must be produced.  In that paper, the box is produced by cropping
          and resizing the crop to a standard size.  In the text, we argue that this cropping suppresses the effects of
          scale and slant.  However, it is a poor model of the boxes produced by modern detectors, because it is placed
accurately round the sign.}
	\label{fig:detect_sign}
\end{figure*}

\section{Classifiers and Detectors are Very Different Systems}

The details of the system attacked are important in assessing the threat posed by Evtimov {\em et al.}'s stop signs.
Their process is: acquire image (or video frame); crop to sign; then classify that box.  This process is 
seen in earlier road sign literature, including~\cite{stallkamp2012man,sermanet2011traffic}.  The attack is on the classifier.  There are two classifiers, distinguished by architecture and training details.  LISA-CNN 
consists of three convolutional layers followed by a fully connected layer (~\cite{evtimov2017robust}, p5, c1), trained to classify signs into 17 classes (~\cite{evtimov2017robust}, p4, c2),
using the LISA dataset of US road signs~\cite{mogelmose2012vision}.  The other is a publicly available implementation (from~\cite{evtimovpaperref29}) of
a classifier demonstrated to work well at road signs (in~\cite{sermanet2011traffic}); this is trained on the German Traffic Sign Recognition Benchmark
(\cite{stallkamp2012man}), with US stop signs added.  Both classifiers are accurate (~\cite{evtimov2017robust}, p5, c1).
Each classifier is applied to $32 \times 32$ images (~\cite{evtimov2017robust}, p4, c2).   However, in 
both stationary and drive by tests, the image is cropped and resized (~\cite{evtimov2017robust}, p8, c2). 

An attack on a road sign {\em classifier} is of no particular interest in and of itself, because no application requires classification
of close cropped images of road signs.  An attack on a road sign {\em detector} is an entirely different matter.  We
interpret Evtimov {\em et al.}'s pipeline as a proxy model of a detection system, where the cropping procedure spoofs
the process in a detector that produces bounding boxes.  This is our interpretation of the paper, but it is not an
unreasonable interpretation; for example, table VII of~\cite{evtimov2017robust} shows boxes placed over small road signs
in large images, which suggests authors have some form of detection process in mind.  We speculate that several features
of this proxy model make it a relatively poor model of a modern detection system.    These features also make the
classifier that labels boxes relatively vulnerable to adversarial constructions.   

The key feature of detection systems is that they tend not to get the boxes exactly right (for example, look at the boxes in
Figure~\ref{fig:detect_sign}), because it is extremely difficult to do. Localization of boxes is measured using the
intersection over union score; one computes $A_I/A_U$, where $A_I$ is the area of intersection between predicted and
true box, and $A_U$ is the area of the union of these boxes. For example, YOLO has a mean Average Precision of 78.6\% at
an IOU score of .5 -- this means that only boxes with IOU with respect to ground truth of .5 or greater are counted as a
true detection.  Even with very strong modern detectors, scores fall fast with increasing IOU threshold.
How detection systems predict boxes depends somewhat on the architecture.  Faster RCNN predicts interesting boxes, then classifies them~\cite{ren2015faster}.
YOLO uses a grid of cells, where each cell uses features computed from much of the image to predict boxes and labels near that cell, with
confidence information~\cite{redmon2016yolo9000}.  One should think of this architecture as an efficient way of
predicting interesting boxes, then classifying them.  All this means that, in modern detector systems, boxes are not
centered cleanly on objects.  We are not aware of any literature on the statistics of box locations with respect to a root
coordinate system for the detected object.

There are several reasons that Evtimov {\em et al.}'s attack on a classifier makes a poor proxy of a detection system.

{\bf Close cropping can remove scale and translation effects:}  The details of the crop and resize procedure are not
revealed in~\cite{evtimov2017robust}.  However, these details matter.  We believe their results are most  easily
explained by assuming the sign was cropped reasonably accurately to its bounding box, then resized (Table VII
of~\cite{evtimov2017robust}, shown for the reader's convenience here as Figure~\ref{fig:detect_sign}).  If the sign is cropped
reasonably accurately to its bounding box, then resized, the visual effects of slant and scale are largely removed.  
In particular, isotropic resizing removes effects of scale other than loss of spatial precision in the sampling grid
This means the claim that the adversarial construction is invariant to slant and scale is moot.   Close  cropping is
not a feature of modern detection systems, and would make the proxy model poor. 

{\bf Low resolution boxes:}  Almost every pixel in an accurately cropped box will testify to the presence of a stop sign.  Thus,
very low resolution boxes may mean that fewer pixels need to be modified to confuse the underlying classifier.  In contrast to the 
32x32 boxes of~\cite{evtimov2017robust}, YOLO uses a 7x7 grid on a 448x448 dimension image; each grid cell predicts
predict bounding box extents and labels.   This means that each prediction in YOLO observes at least  64x64 pixels.
The relatively low resolution of the classifier used makes the proxy model poor. 

{\bf Cropping and variance:}   Detection systems like YOLO or Faster RCNN cannot currently produce accurate bounding
boxes.    Producing very accurate boxes requires searching a larger space of boxes, and so creates  problems with false
positives.  While there are post-processing methods to improve boxes~\cite{gidaris2015locnet}, this  tension is
fundamental (for example, see figure~\ref{fig:detect_imageyolo} and~\ref{fig:detect_imagefrcnn}).   In turn, this means
that the classification procedure within the detector must cope with a range of shifts between box and object.   We
speculate that, in a detection system, this could serve to disrupt adversarial patterns, because the  pattern might be
presented to the classification process inside the detector in a variety of locations relative to the bounding  box.  In
other words, the adversarial property of the pattern would need to be robust to shifts and rescales within the  box.  At
the very least, this effect means that one cannot draw conclusions from the experiments of~\cite{evtimov2017robust}. 

{\bf Cropping and context:} The relatively high variance of bounding boxes around objects in detector systems has
another effect.  The detector system sees object context information that may have been hidden in the  proxy model.  For
example, cells in YOLO do not distinguish between pixels covered by a box and others when deciding  (a) where the box is
and (b) what is in it.   While the value of this information remains moot, its absence means the proxy model is a poor
model.   

\section{Discussion}

We do not claim that detectors are necessarily immune to physical adversarial examples.  Instead, we claim that there is
no evidence as of writing that a physical adversarial example can be constructed that fools a detector.  In earlier
work, we said we had not produced such examples.  The main point of this paper is to point out that others  have not,
too; and that fooling a detector is a very different business from fooling a classifier.   

There is a tension between the test-time accuracy of a classifier, and the ability 
to construct adversarial examples that are ``like'' and ``close to'' real images but are misclassified.  In particular, if there are lots of such things, why is the
classifier accurate on test?  How does the test procedure ``know'' not to produce adversarial examples?  The usual, and natural, explanation is that the measure
of the space of adversarial examples ${\cal A}$ under the distribution of images $P(I)$ is ``small''.  Notice that ${\cal A}$ is interesting only if 
$P({\cal A})$ is small {\em and} for most $u \in {\cal A}$, $P(u)$ is ``big''  (i.e. there is not much point in an adversarial example that doesn't look like an image)  {\em and} there there is at least some of ${\cal A}$ ``far'' from true classifier
boundaries (i.e. there is not much point in replacing a stop sign with a yield sign, then complaining it is mislabelled).  This means that  ${\cal A}$ must have small volume, too.  If ${\cal A}$ has small volume,  but it is easy for an optimization process to find an adversarial example close to any particular example, then there must also be a piece of ${\cal A}$ quite close to most examples (one can think of ``bubbles'' or ``tubes'' of bad labels threading through the space of images).  In this view, Evtimov {\em et al.}'s paper presents an important puzzle.  If one can construct  an
adversarial pattern that remains adversarial for a three dimensional range of views (two angles and a scale), this implies that close to any particular pattern
there is a three parameter ``sheet'' inside ${\cal A}$ -- but how does the network know to organize its errors into a form that is consistent with nuisance viewing parameters?

One answer is that it is trained to do so because it is trained on different views of objects, meaning that ${\cal A}$
has internal structure learned from training examples.  While this can't be disproved, it certainly hasn't been  proved.
This answer would imply that, in some way, the architecture of the network can 
generalize across viewing parameters better than it generalizes across labels (after all, the existence of an adversarial example is a failure to generalize labels correctly).  Believing this requires fairly compelling evidence.   Ockham's razor suggests another answer: Evtimov {\em et al.}, by cropping closely to the stop  sign, removed most of the effect of slant and scale, and so the issue does not arise.

Whether physical adversarial examples exist that fool a detector is a question of the first importance. Here are quite
good reasons they might not.  An adversarial pattern on a physical object that could fool a detector would have to be
adversarial in the face of a wide family of parametric distortions (scale; view angle; box shift inside the detector;
illumination; and so on).  While it is quite possible that the box created by the detector reduces the effects of view
angle and scaling, at least for plane objects, the box shift is an important effect. There is no evidence that
adversarial patterns exist that can fool a detector.  Finding such patterns (or disproving their existence) is an
important technical challenge.  More likely to exist, but  significantly less
of a nuisance,  is  a pattern that, viewed under the right circumstances (and so just occasionally) would fool a  detector.

\subsection*{Acknowledgements}
We are particularly grateful to Ivan Evtimov, Kevin Eykholt, Earlence Fernandes, Tadayoshi Kohno,
Bo Li, Atul Prakash, Amir Rahmati, and Dawn Song, the authors of~\cite{evtimov2017robust}, who have generously shared
data and have made comments on this manuscript which lead to improvements.

{\small
\bibliographystyle{ieee}
\bibliography{egbib}
}

\end{document}